\documentclass[10pt,twocolumn,letterpaper]{article}
\usepackage[table]{xcolor}
\usepackage{cvpr}
\usepackage{graphicx}
\usepackage{amsmath}
\usepackage{amssymb}
\usepackage{booktabs}
\usepackage{colortbl}
\usepackage{multirow}

\definecolor{cvprblue}{rgb}{0.21,0.49,0.74}
\usepackage[pagebackref,breaklinks,colorlinks,citecolor=cvprblue]{hyperref}

\newcommand{\Ours}{CamCtrl3D}

\title{\Ours{}: Single-Image Scene Exploration with Precise 3D Camera Control}
\author{
Stefan Popov \\\vspace{-12pt}\and
Amit Raj \and
Michael Krainin \and
Yuanzhen Li \and
William T. Freeman \and
Michael Rubinstein \\[1.5em] Google DeepMind
}

\begin{document}

\definecolor{olivegreen}{RGB}{0,170,0}
\newcommand{\mypar}[1]{\paragraph{\textbf{#1}}}
\definecolor{bestcol}{RGB}{254,196,79}
\newcommand{\best}[1]{\textbf{#1}}
\definecolor{secondbestcol}{RGB}{255,247,188}
\newcommand{\secondbest}[1]{#1}

\twocolumn[{%
\renewcommand\twocolumn[1][]{#1}%
\maketitle
\vspace{-2.5em}
\begin{center}
    \centering
    \captionsetup{type=figure}
    \includegraphics{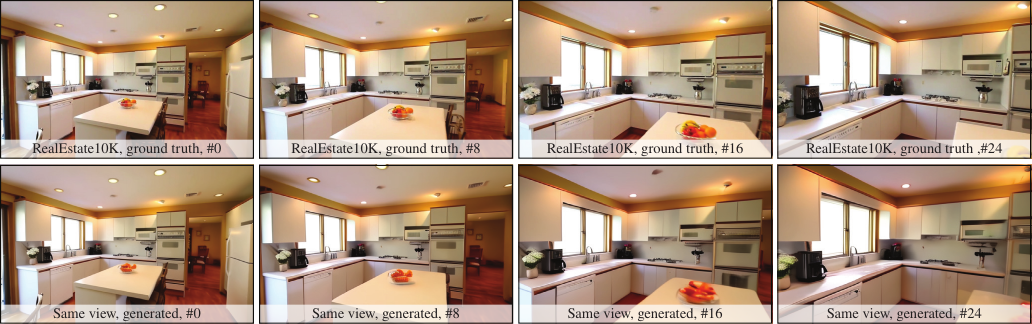}
    \captionof{figure}{
      Our method \Ours{} generates videos of scene fly-throughs, given an
      initial image for frame \#0 and a 3D camera trajectory (bottom row). The
      generated videos are high-quality and closely match the ground truth (top
      row).
      \vspace{.5em}
    }
    \label{fig:teaser}
\end{center}%
}]

\begin{abstract}
    \vspace{-1em}
    We propose a method for generating fly-through videos of a scene,
    from a single image and a given camera trajectory.
    We build upon an image-to-video latent diffusion
    model~\cite{blattmann23svd}. We condition its UNet~\cite{ronneberger15unet}
    denoiser on the camera trajectory, using four techniques.
    (1) We condition the UNet's temporal blocks on raw camera extrinsics, similar to
    MotionCtrl~\cite{wang2023motionctrl}.
    (2) We use images containing camera rays and directions, similar to
    CameraCtrl~\cite{he24cameractr}.
    (3) We re-project the initial image to subsequent frames and use the
    resulting video as a condition.
    (4) We use 2D~$\Leftrightarrow$~3D
    transformers~\cite{tyszkiewicz22raytran} to introduce a global 3D
    representation, which implicitly conditions on the camera poses.
    We combine all conditions in a ContolNet-style~\cite{zhang2023controlnet}
    architecture.
    We then propose a metric that evaluates overall video quality and the
    ability to preserve details with view changes, which we use to analyze the
    trade-offs of individual and combined conditions.
    Finally, we identify an optimal combination of conditions. We calibrate
    camera positions in our datasets for scale consistency across scenes, and we
    train our scene exploration model, \Ours{}, demonstrating state-of-the-art
    results.
\end{abstract}

\section{Introduction}

Generating fly-through videos of a scene from a single image and a predefined
camera trajectory has been a longstanding challenge in the fields of computer
graphics and computer vision. The ultimate goal is to provide users the ability
to walk into their own photographs; to turn a single, specific view of a scene
into a full, immersive viewing experience with minimal capturing effort.

Recent advances in image and video generation
techniques~\cite{ho20ddpm,blattmann23svd,barron22mipnerf,wang2023motionctrl},
have brought us closer to realizing this goal.
In this work, we present an approach that integrates precise 3D camera controls
directly into a pre-trained generative video model. Our approach leverages the
priors learned by the video model to generate realistic and controllable
explorations of a scene captured in a single image.

Several recent works have explored incorporating camera control into existing
video models using indirect conditioning signals, such as raw camera
extrinsics~\cite{wang2023motionctrl} or images with camera ray
coordinates~\cite{he24cameractr,watson244dim}. We adopt these two signals and
propose two additional, novel approaches:
(1) integrating a global 3D representation into the video generation model,
using a physically accurate 2D~$\Leftrightarrow$~3D feature exchange mechanism
(Section~\ref{sec:cond-raytran}), and (2) re-projecting the initial image over
subsequent frames and using the resulting video as a conditioning signal
(Section~\ref{sec:cond-reproj}).
The first approach introduces explicit 3D understanding in the model and enables
inter-frame interactions that are consistent with principles of light transport.
This implicitly conditions the model on the 3D camera poses.
The second approach generates re-projected sequences that closely resemble the
ground truth for surfaces observed in the initial image, allowing the network to
efficiently copy these regions with minimal modification.

We implement these four conditioning approaches (raw camera extrinsics, camera
rays, 2D~$\Leftrightarrow$~3D transformer, initial image reprojection) into a
unified framework and propose a ControlNet-style approach for their combination
(Section~\ref{sec:cnet-cond}). To identify the optimal combination, we examine
the trade-offs of individual and combined conditions, using a
dataset~\cite{dai17scannet} with precise metric-scale camera poses (Section
\ref{sec:results-ablations}). For precise evaluation, we introduce a metric that
considers both the overall quality of the generated videos and the model's
ability to accurately preserve input image details during view changes
(Section~\ref{sec:results-eval-metric}).

Finally, we use the identified optimal conditioning combination (substantial
weight given to  camera extrinsics, 2D~$\Leftrightarrow$~3D, and initial image
re-projection; small weight, albeit still important and improving results
quality, given to camera rays), and we train our scene exploration video model
\Ours{} (Section~\ref{sec:results-full}). We use two datasets that offer crisp
videos with natural framing and diverse content.
The camera poses in these datasets are estimated with
structure-from-motion~\cite{schoenberger16sfm}, and are thus precise only up to
an unknown per-scene global scaling factor.
Thus, to ensure accurate interpretation of scale during camera movement, we
calibrate both datasets to metric scales, using a contemporary metric depth
estimation method (Section~\ref{sec:result-datasets}).

In summary, our contributions are:
(1) We propose two novel camera conditioning techniques based on principles of
light transport;
(2) We integrate these with techniques from existing
works~\cite{wang2023motionctrl,he24cameractr,watson244dim} into a unified
framework;  we analyze the trade-offs of individual and combined conditions and
propose an optimal combination, and then train a scene exploration model \Ours{}
with the optimal combination of conditioning strategies, demonstrating
state-of-the-art results.
(3) We propose a precise metric that evaluates both overall quality and ability
to preserve details with view changes. We then calibrate camera positions in our
datasets, enabling models to interpret scales correctly.

\section{Related work}

\mypar{Novel view synthesis}
Gaussian splatting and NeRF-based
methods~\cite{mildenhall2020nerf,barron22mipnerf,barron23zipnerf,kerbl23gaussian}
achieve high quality novel view synthesis, but require a large number of images
as input and are often trained on a per-scene basis.
SparseFusion~\cite{zhou23sparsefusion} and ReconFusion~\cite{wu23reconfusion}
combine NeRF with diffusion model priors to reduce the required input images,
but still need more than one.
While these methods can generate fly-through videos, they all require more than
one image as input, and significantly more for non-object-centric cases.
Two recent works, CAT3D~\cite{gao24cat3d} and 4DiM~\cite{watson244dim},
demonstrate impressive novel view synthesis results from as few as a single
image, but require extensive training data. CAT3D is trained $\approx$1M posed
videos, while 4DiM is trained on 30M unposed videos and $\approx$250K posed
ones. In contrast, our model is trained on just 10K posed videos.
We quantitatively benchmark our method against 4DiM in
Section~\ref{sec:results-full}, utilizing their reported FVD and PSNR metrics on
the RealEstate10K dataset~\cite{zhou18re10k}.
We do not compare to CAT3D, due to the absence of both single-image quantitative
evaluation results and publicly available source code for this method.

\vspace{-1em}
\mypar{Video models as priors}
The success of diffusion models in image
generation~\cite{ho20ddpm,peebles23diff-trasnf,song20ddim} has inspired a wave
of recent research on video
generation~\cite{he22latent,ho2022videodiff,rombach22ldm,zhang23i2vgen,blattmann23svd,gupta23walt,bartal24lumiere},
both from textual prompts and from single images.
Our method builds on one of these works, namely Stable Video
Diffusion~\cite{blattmann23svd} (SVD).
Most of these methods offer only coarse control over the generated videos,
primarily through the input textual prompts and images.
AnimateDiff~\cite{guo23animatediff} allows transferring motion between videos,
while VideoComposer~\cite{wang23video-composer} allows control through textual,
spatial, and temporal 2D conditions.
Two recent works, MotionCtrl~\cite{wang2023motionctrl} and
CameraCtrl~\cite{he24cameractr}, condition video models on 3D camera
trajectories. We adopt their conditioning signals in our model
(Sections~\ref{sec:cond-raw-extr} and~\ref{sec:cond-rayod}). We further compare
to MotionCtrl in Sections~\ref{sec:results-ablations}
and~\ref{sec:results-full}. CameraCtrl requires additional textual prompts with
input images which is not needed in \Ours{}

\vspace{-1em}
\mypar{Global 3D representations}
Several recent works propose the use of structured latent 3d representation for
novel-view-synthesis. DeepVoxels~\cite{sitzmann19deepvoxels} extracts a
volumetric representation from several views of a scene, using a voxel-based
encoder-decoder. GenNVS~\cite{chan23genvs} lifts a single input view into a
volumetric latent feature grid, with the help of a neural network.
VQ3D~\cite{sargent23vq3d} relies on depth estimation to build a tri-plane
representation of a scene. RayTran~\cite{tyszkiewicz22raytran} relies on sparse
2D$\Leftrightarrow$3D transformers to build a global 3D representation for both
3D reconstruction and rendering novel views. Our method takes inspiration from
these works and proposes a conditioning approach built using RayTran's
2D~$\Leftrightarrow$~3D sparse transformers.

\section{Proposed Approach}
\label{sec:approach}

\begin{figure*}
    \includegraphics{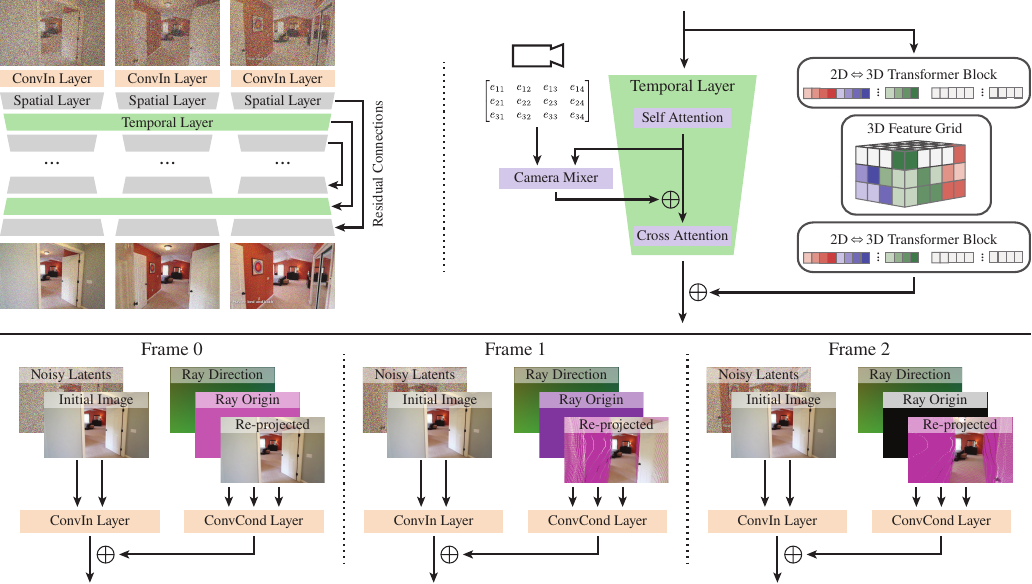}
    \caption{ \textbf{Top left:} We add camera conditioning to the UNet denoiser
    of SVD~\cite{blattmann23svd} by modifying its layers. \textbf{Top right:} We
    attach the camera extrinsics and the 2D$\Leftrightarrow$3D transformer
    conditions to UNet's temporal layers (Sections~\ref{sec:cond-raw-extr} and
    \ref{sec:cond-raytran}). \textbf{Bottom:} We add additional top-level
    convolutional layers for the camera ray and re-projected image conditions
    (Sections~\ref{sec:cond-rayod} and \ref{sec:cond-reproj}) }
    \label{fig:unet-block-cond}
\end{figure*}

\Ours{} takes an initial RGB image $\mathbf{I_0}$ and a sequence of camera poses
$\{\mathbf{c}_i\}_{i=0}^N$ as input. The image depicts a virtual 3D scene
$\mathbf{V}$ from the perspective of the first camera. As output, \Ours{}
generates a sequence of views $\{{\mathbf{I}_i}\}_{i=0}^N$ of the virtual scene
$\mathbf{V}$, corresponding to the remaining cameras.

To achieve this, we modify a pretrained video generation model, Stable Video
Diffusion~\cite{blattmann23svd} (SVD), and more precisely its UNet denoiser. We
condition UNet's temporal blocks on the raw camera extrinsics
(Section~\ref{sec:cond-raw-extr}). We further provide UNet with images
containing the camera ray origin $\mathbf{o}_i \in \mathbb{R}^{W \times H \times
3}$ and directions $\mathbf{d}_i \in \mathbb{R}^{W \times H \times 3}$ for each
frame $i$ (Section~\ref{sec:cond-rayod}). We re-project the input image with the
camera poses using estimated depth and we condition UNet on the resulting video
(Section~\ref{sec:cond-reproj}). We introduce 3D understanding to UNet, using a
global 3D representation and sparse 2D $\Leftrightarrow$ 3D transformer blocks,
and we condition UNet in 3D (Section~\ref{sec:cond-raytran}). Finally, we
combine all conditions in a ControlNet-style~\cite{zhang2023controlnet}
architecture (Section~\ref{sec:cnet-cond}).

\subsection{Preliminaries}
SVD~\cite{blattmann23svd} is a latent video diffusion
model~\cite{blattmann23vldm}, fine-tuned for high-resolution image-to-video
generation. It uses a reverse diffusion~\cite{sohl15diffusion} process with a
learned UNet denoiser to generate videos in a latent representation, and a
variational autoencoder~\cite{blattmann23vldm} to convert to and from an RGB
representation.
UNet has an encoder-decoder architecture with residual connections. It is built
from alternating spatial and temporal blocks. Spatial blocks operate on video
frames independently, across their pixels. Temporal blocks operate across time,
independently within each pixel.
In the following sections, we introduce camera pose conditioning, by modifying
UNet's inputs and its temporal blocks.

\subsection{Condition on raw camera extrinsics}
\label{sec:cond-raw-extr}
Temporal blocks consist of self-attention across time, followed by
cross-attention, with features extracted from the input image using
CLIP~\cite{radford2021clip}. We condition on the raw camera extrinsics, by
inserting a residual block between the two attention layers
(Figure~\ref{fig:unet-block-cond}). In it, we concatenate the 12 entries of the
$4 \times 3$ camera extrinsics matrix to the features of each pixel in each
frame. We then use a feed forward network to compress the features to match the
cross-attention dimensions.
This is similar to MotionCtrl~\cite{wang2023motionctrl}, however we incorporate
the feed forward outputs as residuals to facilitate back propagation.

\subsection{Condition on camera rays}
\label{sec:cond-rayod}
For each frame, we compute two new guiding images: $\mathbf{d}_i$ and
$\mathbf{o}_i$. The first contains the direction of the camera rays passing
through each pixel of frame $i$ in world space coordinates, the second contains
the camera origins, again in world space coordinates.
We normalize $\mathbf{d}_i$'s values to the range $[0, 1]$, by adding 1 and
dividing by 2. To ensure all values in $\mathbf{o}_i$ are positive, we offset
all camera origins into the positive octant (+++) beforehand.

We encode the two resulting videos into a latent representation using SVD's VAE.
We feed the result into a new convolutional layer and add its output to UNet's
first convolutional layer (Figure~\ref{fig:unet-block-cond}). Similar to
Section~\ref{sec:cond-raw-extr}, this conditions the model on the camera
parameters, however this representation is more natural as the model can reason
about camera motion at pixel level.

\begin{figure}
    \includegraphics{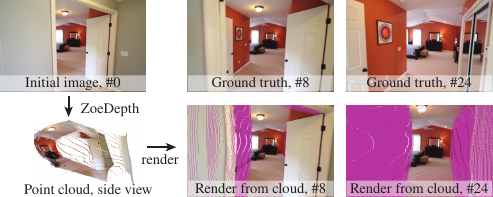}
    \caption{We re-project the surfaces observed on the initial image to all
    subsequent frames, using ZoeDepth~\cite{bhat23zoedepth} to estimate a point
    cloud. We use the resulting frames as a condition
    (Section~\ref{sec:cond-reproj}) and during evaluation
    (Section~\ref{sec:results-eval-metric}).}
    \label{fig:vid-cond}
\end{figure}

\subsection{Condition on re-projected initial image}
\label{sec:cond-reproj}
We re-project the surface observed in the initial image to the rest of the
frames and condition UNet on the resulting video.
To do this, we first apply a metric-space monocular depth estimation model
ZoeDepth ~\cite{bhat23zoedepth} to the input image. We combine the resulting
depth with the pixel colors and unproject using the parameters of the first
camera $\mathbf{c}_0$. This results in a point cloud, which we render onto the
subsequent frames using their respective cameras $\mathbf{c}_i$
(Figure~\ref{fig:vid-cond}).
We encode the resulting video into latent space, we feed the result into a new
convolutional layer, and similar to Section~\ref{sec:cond-rayod} we add the
output to UNet's first convolutional layer (Figure~\ref{fig:unet-block-cond}).

The point cloud captures the visible surface of the input image. Assuming a
static scene, the rendered video tracks this surface consistently with the
camera motion. Additionally, we use a distinct background color during rendering
that is unlikely to occur naturally. This allows the model to both stay
consistent with the initial image and generate new content in place of the
background color.

\subsection{Condition using 2D$\Leftrightarrow$3D transformers}
\label{sec:cond-raytran}

Intrinsically, UNet operates in 2D, on arrays of 2D grids corresponding to each
frame. They are connected through the time dimension, in UNet's temporals
blocks. We propose to supplement these with a new type of block that operates on
a global 3D representation (Figure~\ref{fig:unet-block-cond}).

As input, the block accepts an array of 2D features, as well as their
corresponding camera parameters (intrinsic and extrinsic camera matrices).
We use a voxel grid of features as a 3D representation. The grid has fixed
dimensions
and it is centered \wrt the camera origins. Its resolution varies, depending on
where the block is placed inside UNet. We use sparse ray-traced
attention~\cite{tyszkiewicz22raytran} to project the input 2D array into the
voxel grid, with the given camera parameters. We then use a convolutional 3D
encoder-decoder with residual connections to enable reasoning in 3D. Finally, we
project back onto the 2D array using sparse ray-traced attention once more.
We also embed time into the feature vectors before a 2D $\Rightarrow$ 3D
projection, using positional encoding. This allows reasoning across time in the
3D representation, thus enabling dynamic scenes.
We use the new 3D blocks alongside UNet's temporal blocks, and we add their
outputs.

Ray-traced attention embeds knowledge about the image formation process directly
into the model. It allows the network to jointly analyze all views and to
consolidate the extracted information into a global 3D representation. It is
known to work well for 3D reconstruction from RGB
videos~\cite{tyszkiewicz22raytran, rukhovich22imvoxnet}, as well as for view
interpolation~\cite[supplementary material]{tyszkiewicz22raytran}.
In our case, ray-traced attention allows the network to reason about the world
contents directly in 3D, and to then project this into the individual video
frames.

\subsection{Combine conditions with ControlNet}

\begin{figure}
    \centering
    \includegraphics{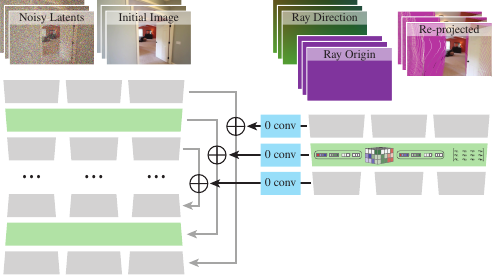}
    \caption{We apply conditions to a clone of the UNet encoder
    (Section~\ref{sec:cnet-cond}), and we add its outgoing residual connections
    to those of the original encoder, after passing through zero
    convolutions~\cite{zhang2023controlnet}. }
    \label{fig:cnet-cond}
\end{figure}

\label{sec:cnet-cond}
We incorporate the above conditions into UNet in a
ControlNet-style~\cite{zhang2023controlnet} architecture. We clone UNet's
encoder and we attach all conditioning layers to it. We attach zero convolution
layers~\cite{zhang2023controlnet} to its outgoing residual connections and we
add their outputs to the respective residual connections in the original encoder
(Figure~\ref{fig:cnet-cond}).

\section{Experiments}

We first perform a set of ablations to study the trade-offs of the different
conditioning methods (Section~\ref{sec:results-ablations}). Building on insights
from this study, we develop an optimal conditioning strategy and we train a
final high-quality video model \Ours{} (Section~\ref{sec:results-full}).

\subsection{Evaluation metric}
\label{sec:results-eval-metric}
\begin{figure}
    \centering
    \includegraphics{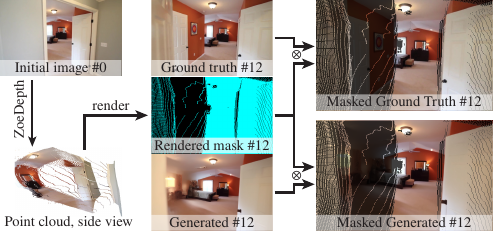}
    \caption{Re-projection (Sec.~\ref{sec:cond-reproj}) identifies regions
    within a frame originating from the initial image (\eg frame \#12 here). We
    apply the resulting mask to both ground truth and generated images and
    measure image difference (Section~\ref{sec:results-eval-metric}) to assess the model's ability to maintain visual consistency during camera change. }
    \label{fig:eval-mask}
\end{figure}

To assess the quality of the generated videos, we consider two key factors: (1)
the overall quality of the generated videos, and (2) the model's ability to
maintain details from the input image as the view changes.

For (1), we compare the distribution of the generated videos to that of the test
set, using Fréchet Video Distance~\cite{unterthiner19fvd}  (FVD).
For (2), we assume a static scene and known pixel depths in the first frame.
Similar to Section~\ref{sec:cond-reproj}, we re-project the first frame onto
each subsequent frame, using the provided camera poses. This creates a binary
mask per frame, identifying pixels that originate from the first frame versus
those with new content (Figure~\ref{fig:eval-mask}). We use this mask to compare
the generated video against ground truth, ensuring masked pixels match exactly.
We measure peak signal-to-noise ratio (PSNR) for the difference of these pixels,
along with LPIPS~\cite{zhang18lpips} and SSIM~\cite{wang04ssim}. In some cases,
we also report peak signal-to-noise ratio computed on the full video frame
(FPSNR), disregarding the mask.

\subsection{Datasets}
\label{sec:result-datasets}

\begin{figure}
    \centering
    \includegraphics{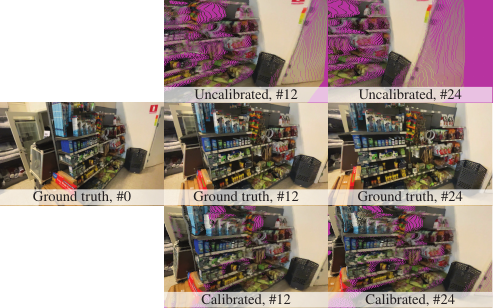}
    \caption{ Metric-calibration on DL3DV, frames \#12 and \#24. Uncalibrated
    re-projections (top row) deviate significantly from the ground truth (middle
    row), hindering both the re-projected condition
    (Section~\ref{sec:cond-reproj}) and evaluation
    (Section~\ref{sec:results-eval-metric}). Calibration (bottom row) rectifies
    this discrepancy. }
    \label{fig:eval-dataset-calibration}
\end{figure}

We use three datasets in our experiments: ScanNet~\cite{dai17scannet},
RealEstate10K~\cite{zhou18re10k}, and DL3DV~\cite{ling24dl3dv}.
ScanNet contains videos of indoor spaces, captured with an RGB-D sensor. It
offers ground-truth depth maps and precise camera poses. Because of this, we use
it in our ablation studies, training on 1194 videos, and evaluating on 312.
RealEstate10K and DL3DV contain videos of indoor and outdoor spaces. They offer
crisp videos with natural framing and diverse content. We use them for our final
model, training on 4937 RealEstate10K and 4497 DL3DV videos, and evaluating on
312 different videos from each dataset respectively.

\mypar{Sampling clips}
The videos in these datasets are much longer than our models' output. Therefore,
we sample clips matching that length, at varying sampling speeds. During
training, we choose a random starting frame $f$ and a random fractional sampling
speed $s$, between 1 and 10. We take the $F$ frames with indices $\lfloor f +i s
\rfloor$, where $F$ is the output length of our model, and $i \in [0, F)$ is an
integer. Conversely, during evaluation we always start from the first frame and
we use a fixed sampling speed of either 1, 2, 4, or 8.

Varying the sampling speed offers control over camera motion during both
training and evaluation. Furthermore, it influences the ratio of pixels observed
on the initial image versus newly generated pixels within a frame. For the test
split of ScanNet, this ratio increases proportionally with sampling speed: 4.9\%
at speed $\times$1, 10.5\% at speed $\times$2, 21.8\% at speed $\times$4, and
36.7\% at speed $\times$8.

\begin{table*}
    \centering
\begin{minipage}{\linewidth}
\resizebox{\columnwidth}{!}{
    \setlength{\tabcolsep}{3pt}
    \begin{tabular}{l rrrr rrrr rrrr rrrr}
    \toprule
  & \multicolumn{4}{c}{Speed $\times$1} & \multicolumn{4}{c}{Speed $\times$2 } & \multicolumn{4}{c}{Speed $\times$4 } & \multicolumn{4}{c}{Speed $\times$8 }\\[.2em]
  \cmidrule(lr){2-5}
  \cmidrule(lr){6-9}
  \cmidrule(lr){10-13}
    \cmidrule(lr){14-17}
 Condition &{\small FVD$\downarrow$}&{\small PSNR$\uparrow$}&{\small SSIM$\uparrow$}&{\small LPIPS$\downarrow$}&{\small FVD$\downarrow$}&{\small PSNR$\uparrow$}&{\small SSIM$\uparrow$}&{\small LPIPS$\downarrow$}&{\small FVD$\downarrow$}&{\small PSNR$\uparrow$}&{\small SSIM$\uparrow$}&{\small LPIPS$\downarrow$}&{\small FVD$\downarrow$}&{\small PSNR$\uparrow$}&{\small SSIM$\uparrow$}&{\small LPIPS$\downarrow$}\\[.1em]
\midrule
Baseline&218.4&13.0&0.49&0.41&149.7&13.3&0.52&0.38&163.0&13.7&0.58&0.34&303.6&14.2&0.65&0.28\\
\hline
$X$&172.1&13.9&0.51&0.38&132.2&14.0&0.53&0.36&158.9&14.4&0.59&0.32&248.2&14.7&0.66&0.27\\
$C$&78.3&17.1&0.59&0.28&106.8&16.8&0.60&0.28&138.4&16.1&0.63&0.28&172.8&16.0&0.69&0.24\\
$P$&47.3&23.9&0.79&0.15&71.5&22.7&0.78&0.16&\best{107.2}&21.4&0.77&0.18&146.6&20.3&0.78&0.18\\
$R_1$&51.5&21.8&0.73&0.18&76.3&20.9&0.72&0.19&118.0&20.0&0.73&0.20&149.8&19.2&0.76&0.19\\
$R_2$&47.9&22.4&0.75&0.17&73.1&21.3&0.74&0.18&116.3&20.3&0.74&0.19&152.3&19.5&0.76&0.19\\
$R_3$&49.0&22.2&0.74&0.17&73.5&21.2&0.74&0.18&117.4&20.3&0.74&0.19&146.9&19.4&0.76&0.19\\
$R_4$&50.6&21.9&0.73&0.18&74.8&20.9&0.72&0.19&113.8&20.0&0.73&0.20&148.5&19.3&0.76&0.19\\
\hline
$C+R_2$&49.5&22.2&0.74&0.17&73.6&21.1&0.73&0.18&110.1&20.2&0.74&0.19&145.7&19.3&0.76&0.19\\
$X+R_2$&48.6&22.8&0.76&0.16&72.7&21.6&0.75&0.18&111.2&20.5&0.75&0.19&150.6&19.5&0.77&0.18\\
$X+R_2+P$&44.4&\best{24.8}&\best{0.82}&\best{0.14}&68.3&23.3&\best{0.80}&\best{0.15}&109.2&21.9&\best{0.79}&\best{0.17}&\best{143.3}&20.9&\best{0.80}&\best{0.17}\\
$X+R_2+P+C$&\best{42.6}&\best{24.8}&\best{0.82}&\best{0.14}&\best{68.2}&\best{23.4}&\best{0.80}&\best{0.15}&108.8&\best{22.1}&\best{0.79}&\best{0.17}&143.7&\best{21.0}&\best{0.80}&\best{0.17}\\
\hline
MotionCtrl~\cite{wang2023motionctrl}&221.6&14.1&0.44&0.41&198.9&11.2&0.47&0.11&176.6&16.1&0.61&0.67&252.4&13.3&0.63&0.34\\
\bottomrule

\end{tabular}
}
\end{minipage}
    \caption{ Performance of our ablation models from
Section~\ref{sec:results-ablations} and MotionCtrl~\cite{wang2023motionctrl} on
the test set of ScanNet, at different sampling speeds. The best result for each
metric is highlighted in bold. $X$ denotes raw extrinsics conditioning
(Section~\ref{sec:cond-raw-extr}), $C$ denotes camera rays
(Section~\ref{sec:cond-rayod}), $P$ denotes initial image re-projection
(Section~\ref{sec:cond-reproj}), and $R_x$ denotes 2D$\Leftrightarrow$3D
transformers (Section~\ref{sec:cond-raytran}) attached to $x$ UNet layers. We
use an SVD model fine-tuned over ScanNet as baseline. See text for more detail.
}
    \label{tab:ablation-study}
\end{table*}

\mypar{Metric calibration}
The camera poses in RealEstate10K and DL3DV are estimated with
structure-from-motion~\cite{schoenberger16sfm} (SfM), and are thus precise only
up to an unknown per-scene global scaling factor.
This is problematic when conditioning on a single image, as the model cannot
learn the meaning of scale in the user-provided input camera path. Moreover, the
video's motion becomes inconsistent with the re-projected motion described in
Section~\ref{sec:cond-reproj}, making this conditioning approach inappropriate
(Figure~\ref{fig:eval-dataset-calibration}).

We thus calibrate the two datasets. For each frame, we first estimate a
metric-scale depth map, using ZoeDepth~\cite{bhat23zoedepth}. We then project
the SfM point cloud onto the frame. For each SfM point, the ratio of its camera
depth to the depth provided by ZoeDepth serves as an estimate of the global
scaling factor for the entire video. We calculate a robust estimate of this
factor by taking the mean of the depth ratios across all points and frames,
after excluding the smallest and largest 10\% of values. We then apply the
global scaling factor to the camera positions for the video, multiplying them
accordingly.
To assess the accuracy of this estimation, we examined 10 random videos from
each dataset, along with the 10 videos exhibiting the highest variability in
per-point scales. In every instance, the observed motion within the videos
closely aligned with the motion of the re-projected first frame from
Section~\ref{sec:cond-reproj}.

\subsection{Ablation studies}
\label{sec:results-ablations}

To investigate the trade-offs of different conditioning techniques, we conduct
ablation studies, starting with individual experiments for each of the methods
from  Section~\ref{sec:approach}: raw extrinsics denoted as $X$ below
(Section~\ref{sec:cond-raw-extr}), camera rays denoted as $C$
(Section~\ref{sec:cond-rayod}), and initial image re-projection denoted as $P$
(Section~\ref{sec:cond-reproj}). For the 2D$\Leftrightarrow$3D transformers
condition (Section~\ref{sec:cond-raytran}), denoted as $R_x$, we additionally
ablate on the number $x$ of UNet blocks that the condition is attached to (1, 2,
3, or 4), as the decreasing resolution of the 2D grids in deeper UNet blocks
could potentially degrade the performance of this conditioning technique. We
then conduct experiments on combinations of conditioning techniques.

In each experiment, we train a model to generate 14-frame videos with a
resolution of 512x320 pixels.
We use the \emph{train} split of ScanNet~\cite{dai17scannet} for training, the
\emph{val} split for evaluation, and we sample clips as described above. We
resize the dataset videos to the model's resolution in an aspect-preseving way,
using center cropping.

Given the resolution difference between our models and the publicly released SVD
model, we first fine-tune the latter on ScanNet videos for 360K steps. We use
the resulting model to initialize training in our ablation studies and we also
benchmark against it. We train all ablation models for 250K steps. Consistent
with ControlNet ~\cite{zhang2023controlnet}, we observe sudden convergence, at
around 25K steps for the $R_x$ and $P$ conditions, while $C$ and $X$ converge
later, at around 60K steps.

The results of our experiments are summarized in Table~\ref{tab:ablation-study}.
All conditioning methods demonstrate an improvement over the baseline. When
evaluated independently, the re-projected image condition $P$ performs best.
This is expected, as significant portions of the condition closely resemble the
ground truth videos, allowing the network to readily incorporate them with
minimal modification. The 2D$\Leftrightarrow$3D transformer conditions $R_x$
follow closely in performance. Among them, attaching to two UNet layers performs
best ($R_2$). Models conditioned directly on raw extrinsic matrices $X$ perform
worst, as this approach requires learning the complex relationship between 3D
extrinsic matrix values and their corresponding 2D image changes across all
frames.

Combining raw camera extrinsics with 2D$\Leftrightarrow$3D transformers linked
to two UNet layers, outperforms either conditioning method alone ($X+R_2$).
Adding initial image re-projection further enhances performance ($X+R_2+P$),
outperforming all individual conditioning methods. Adding camera ray
conditioning yields marginal improvements, resulting in the optimal technique
($X+R_2+P+C$).

We also measure the performance of MotionCtrl~\cite{wang2023motionctrl} on our
test set using our evaluation metric (see Table~\ref{tab:ablation-study}). For a
fair comparison, we maximize MotionCtrl's performance on our test set by tuning
its FPS and motion magnitude parameters. MotionCtrl performs on-par with our
re-implementation $X$. All other conditioning methods outperform it. Our optimal
configuration $X+R_2+P+C$ has $5.2$ times lower FVD, and $10.7$ dB higher PSNR,
at video sampling speed $\times$1.

\subsection{Scene exploration model \Ours{}}
\label{sec:results-full}

\begin{table}[]
    \centering
    \begin{minipage}{\linewidth}
    \resizebox{\columnwidth}{!}{
    \small
    \setlength{\tabcolsep}{3pt}
    \begin{tabular}{rcrrrrr}
Method&Dataset and speed&FVD$\downarrow$&PSNR$\uparrow$&LPIPS$\downarrow$&FPSNR$\uparrow$\\
\hline
MotionCtrl~\cite{wang2023motionctrl}&RealEstate10K $\times$1&777.5&16.1&0.37&15.6\\
4DiM~\cite{watson244dim}&RealEstate10K $\times$1&195.1&-&&18.1\\
\Ours{}&RealEstate10K $\times$1&\best{72.8}&\best{21.4}&\best{0.13}&\best{20.6}\\
\midrule
\Ours{}&RealEstate10K $\times$2&105.1&20.0&0.15&18.6\\
\Ours{}&RealEstate10K $\times$4&152.7&18.9&0.16&16.5\\
\Ours{}&DL3DV $\times$1&245.1&17.7&0.26&15.9\\
\end{tabular}
     }
    \end{minipage}
    \caption{Performance of our final model \Ours{} (Section~\ref{sec:results-full}) and
    MotionCtrl~\cite{wang2023motionctrl}, on the RealEstate10K
    and DL3DV datasets, at different sampling speeds. We also include metrics for 4DiM, as
    reported in ~\cite{watson244dim}. Our model achieves significantly better quality,
    compared to MotionCtrl and 4DiM.
    }
    \label{tab:results-final}
\end{table}

For our final scene exploration model \Ours{}, we employ the optimal combination
of conditioning strategies identified in Section~\ref{sec:results-ablations}. We
maintain a resolution of $512\times320$, while generating 25 frames per
sequence. We initialize our model with the weights from the official 25-frame
SVD model. We use the RealEstate10K and DL3DV datasets, with splits and sampling
strategies as described in Section~\ref{sec:result-datasets}. Rather than
pre-training and freezing as in Section~\ref{sec:results-ablations} we train the
full model, including the original UNet encoder and decoder, for 1.66M steps.

\begin{figure*}[p]
    \centering
    \includegraphics[width=.92\textwidth]{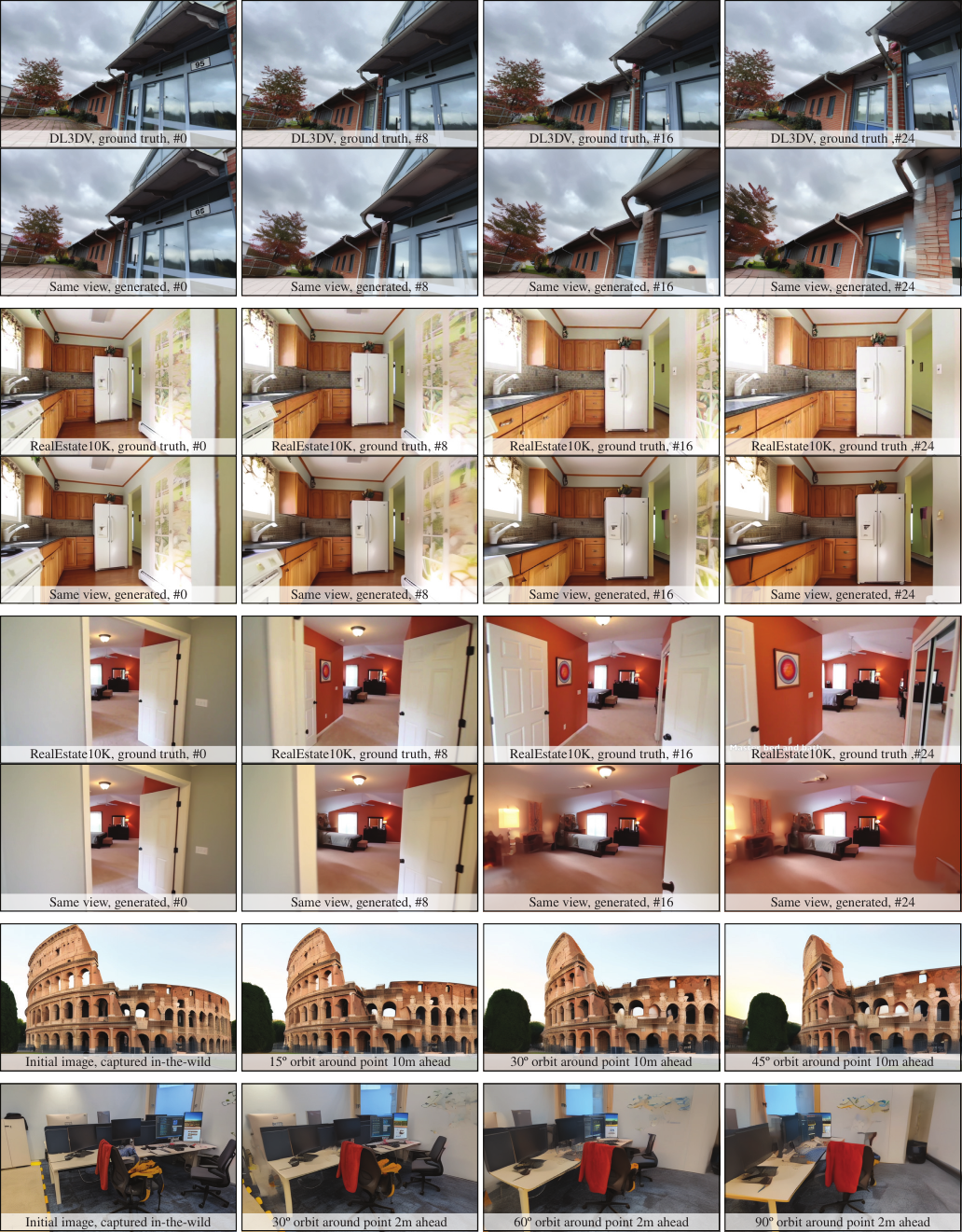}
    \caption{Results generated by the final model (Section~\ref{sec:results-full}).
    In all examples, we show frames \#0, \#8, \#16, and \#24 from the 25 frame
    video. The top 3 examples show videos from the test sets of RealEstate10K
    and DL3DV. Each example contains two rows, one showing ground truth (top) and
    another showing generated results (bottom). The bottom two examples contain video
    sequences generated from images in the wild. Both examples show orbiting
    camera motion. }
    \label{fig:results-qualitative}
\end{figure*}

Figure~\ref{fig:results-qualitative} shows the outputs of our model on the
RealEstate10K and DL3DV test sets, while Table~\ref{tab:results-final} presents
its quantitative evaluation. Our model generates high-fidelity videos with
accurate camera trajectories, even for complex scenes, typical for the
RealEstate10K and DL3DV datasets.

We benchmark our model against two state-of-the-art methods for camera control
in video generation: MotionCtrl~\cite{wang2023motionctrl} and
4DiM~\cite{watson244dim}.
Direct comparison on identical datasets and metrics
(Table~\ref{tab:results-final}) shows our model's significant improvement over
MotionCtrl, achieving up to 8 times reduction in FVD and 4 dB PSNR gain.
Without access to runnable source code for 4DiM, we compare our model's FVD and
FPSNR to their reported values. We use the metric-calibrated test set of
RealEstate10K at $\times$1 sampling speed, since both works report numbers on
it, albeit with different calibration approaches. Our model achieves
significantly better quality, with 2.7$\times$ lower FVD (72.8 \vs 195.1) and
2.6 dB higher FPSNR (20.7 \vs 18.1). At the same time our model requires
significantly less training data (10K posed videos) than 4DiM (30M unposed
videos and $\approx$250K posed videos).

\section{Discussion}

We have shown that by leveraging priors from video models, along with a carefully selected
set of conditioning techniques, \Ours{} can generate fly-throughs of scenes from
a single image. Due to the nature of our task, our training sets primarily
consists of videos of static scenes. Thus, \Ours{} tends to mainly output videos
of static scenes. Occasionally, the model is able to animate parts of them (\eg
waves moving in the ocean) due to the video priors. We observe that the model
relies on the different conditioning techniques to a varying extent based on the
content of the initial image, thus allowing for motion in certain kinds of
scenes. We hypothesize that the model's ability to generate videos with
moving objects can be enhanced through fine-tuning on a dataset comprising
dynamic scenes with calibrated camera parameters.

Further, we observed that models generating 25-frame sequences are better at
maintaining video quality with greater camera motion, compared to models
generating 14-frame sequences. We posit this is likely because they can reason
on smaller inter-frame changes. We experimented with even longer sequences (up
to 80 frames) but observed the reverse effect. We fine tune SVD \emph{with only
10K videos}, and it is likely that the number of training examples does not
provide sufficient information to condition the base model to generate longer
videos.

In conclusion, we introduced two novel camera conditioning techniques based on
light transport principles, and combined these with existing methods within a
unified framework. We trained a method that generates fly-through videos from a
single image and a camera trajectory, with state-of-the-art performance.

{
    \small
    \bibliographystyle{ieeenat_fullname}
    \bibliography{main}
}

\end{document}